\documentclass{article}

\PassOptionsToPackage{numbers, compress}{natbib}

\usepackage[preprint]{neurips_2025}

\usepackage[utf8]{inputenc} 
\usepackage[T1]{fontenc}    
\usepackage{hyperref}       
\usepackage{url}            
\usepackage{booktabs}       
\usepackage{amsfonts}       
\usepackage{nicefrac}       
\usepackage{microtype}      
\usepackage{xcolor}         

\usepackage{graphicx}     
\usepackage{tabularx}    
\usepackage{booktabs}     
\usepackage{multirow}    
\usepackage{amssymb}     
\usepackage{pifont}      
\usepackage{amssymb}
\usepackage{xcolor} 
\usepackage{tikz}

\usepackage{amsmath}

\usepackage{makecell} 

\usepackage{wrapfig}

\AtEndPreamble{
    \usepackage[capitalize]{cleveref}
    \crefname{section}{Sec.}{Secs.}
    \Crefname{section}{Section}{Sections}
    \Crefname{table}{Table}{Tables}
    \crefname{table}{Tab.}{Tabs.}
}

\title{A Preliminary Study on GPT-Image Generation Model for Image Restoration}

\author{%
  Hao Yang$^1$, Yan Yang$^2$, Ruikun Zhang$^1$, Liyuan Pan$^1$ \\
  $^1$Beijing Institute of Technology, $^2$Australian National University \\
  \texttt{hao.yang@bit.edu.cn} \\
  Our GPT-restored Results are publicly available at \\ \url{https://github.com/noxsine/GPT_Restoration}
}

\begin{document}
\maketitle


\begin{abstract}
Recent advances in OpenAI's GPT-series multimodal generation models have shown remarkable capabilities in producing visually compelling images. In this work, we investigate its potential impact on the image restoration community. We provide, to the best of our knowledge, the first systematic benchmark across diverse restoration scenarios. Our evaluation shows that, while the restoration results generated by GPT-Image models are often perceptually pleasant, they tend to lack pixel-level structural fidelity compared with ground-truth references. Typical deviations include changes in image geometry, object positions or counts, and even modifications in perspective. Beyond empirical observations, we further demonstrate that outputs from GPT-Image models can act as strong visual priors, offering notable performance improvements for existing restoration networks. Using dehazing, deraining, and low-light enhancement as representative case studies, we show that integrating GPT-generated priors significantly boosts restoration quality. This study not only provides practical insights and a baseline framework for incorporating GPT-based generative priors into restoration pipelines, but also highlights new opportunities for bridging image generation models and restoration tasks. To support future research, we will release GPT-restored results.
\end{abstract}

\section{Introduction}
\begin{figure}[!t]

    \centering
    \begin{tikzpicture}
    \centering
    \node[anchor=south west,inner sep=0] (image) at (0,0) {\includegraphics[width=0.99\linewidth]{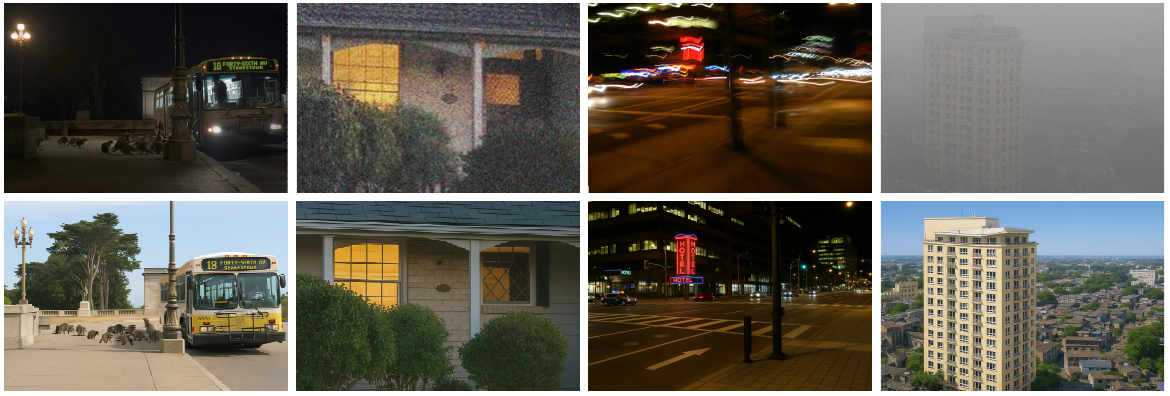}};
    \begin{scope}[x={(image.south east)},y={(image.north west)}]

    \draw (0.12,-0.05) node {\small (a)};
    \draw (0.38,-0.05) node {\small (b)};
    \draw (0.63,-0.05) node {\small (c)};
    \draw (0.88,-0.05) node {\small (d)};

    \end{scope}
    \end{tikzpicture}

    \caption{Image restoration results of GPT-Image on real world degradation without available ground truth. The first row and second row are degraded inputs and the restored outputs, respectively. (a)-(d) correspond to low-light conditions, heavy noise, motion blur, and dense haze, respectively.}
    \label{fig:intro}

\end{figure}
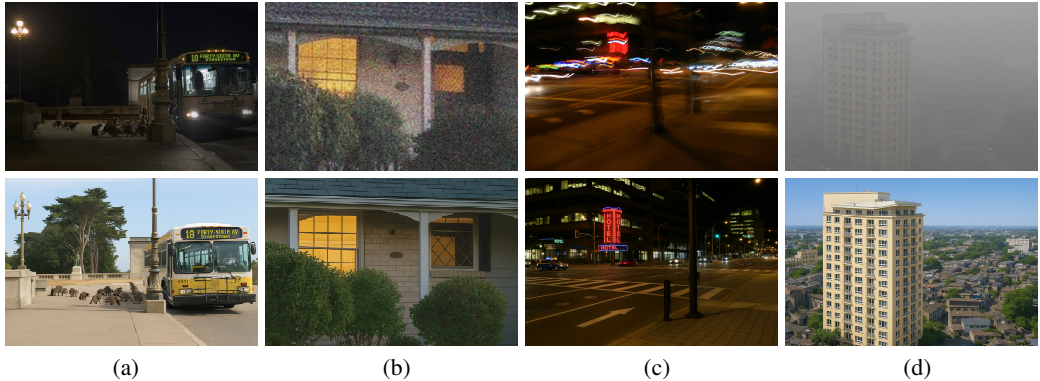

Multimodal large language models have made groundbreaking progress in visual generation \cite{yan2025gpt}. Among them, OpenAI’s GPT-Image (an official image generation model released in the GPT-Image  model era) \cite{openai_gpt_image_1} stands out for its ability to interpret complex visual and textual inputs to generate semantically accurate, visually realistic images. Meanwhile, image restoration can be naturally formulated as a conditional image generation task \cite{croitoru2023diffusion}, where degraded images serve as visual conditioning inputs. By providing an appropriate prompt, GPT-Image's generative capabilities can be directed toward image restoration. This capability represents a major leap forward in multimodal generation and has prompted renewed consideration of its role in image restoration tasks (see \cref{fig:intro}).

Traditionally, image restoration methods rely on degradation-specific network architectures designed to achieve high performance on individual tasks, such as image denoising \cite{abdelhamed2018high}, image deblurring \cite{abuolaim2020defocus}, image super-resolution \cite{wu2024ultralight}, image deraining \cite{zhang2019image_rain800}, and image dehazing \cite{Liu2025Resid}. While these methods are effective within their respective domains, they often lack flexibility and exhibit poor generalization across diverse degradation types. Although some recent efforts have explored unified, all-in-one frameworks capable of handling multiple restoration tasks within a single model \cite{jiang2024survey}, such approaches have yet to demonstrate scalability or consistent performance across varied restoration scenarios.

Given its powerful visual generation and semantic understanding capabilities, GPT-Image naturally emerges as a potential foundation model for all-in-one image restoration \cite{jiang2024survey}. In this work, we conduct the first systematic investigation of GPT-Image in the context of image restoration, uncovering both its promising strengths and current limitations. Building on these insights, we further explore a simple baseline approach that leverages GPT-Image as a plug-and-play component to enhance the performance of existing restoration networks. The study is organized into three parts.

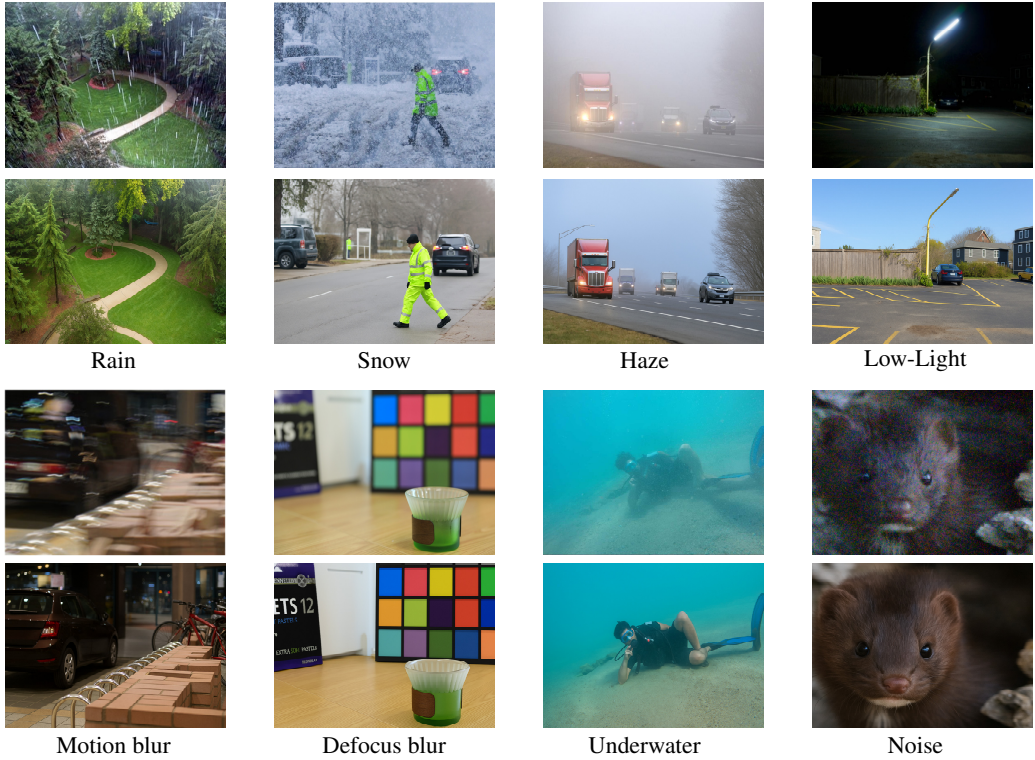
\begin{figure*}[!t]
    \centering

    \begin{tikzpicture}
    \centering
    \node[anchor=south west,inner sep=0] (image) at (0,0) {\includegraphics[width=0.99\linewidth]{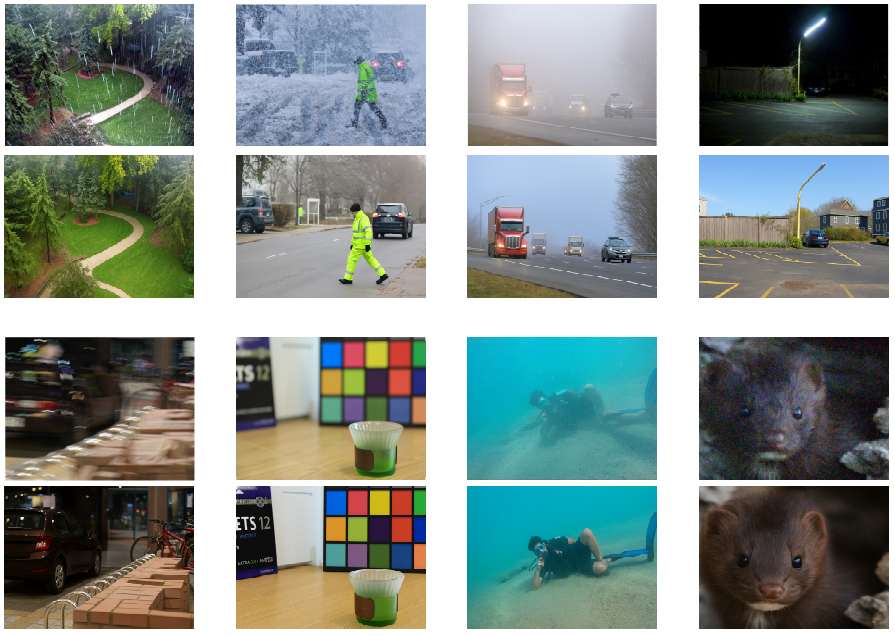}};
    \begin{scope}[x={(image.south east)},y={(image.north west)}]
    
    \draw (0.11,0.51) node{\footnotesize Rain};
    \draw (0.37,0.51) node{\footnotesize Snow};
    \draw (0.62,0.51) node{\footnotesize Haze};
     \draw (0.88,0.51) node{\footnotesize Low-Light};
    \draw (0.11,-0.01) node{\footnotesize Motion blur};
    \draw (0.37,-0.01) node{\footnotesize Defocus blur};
    \draw (0.62,-0.01) node{\footnotesize Underwater};
     \draw (0.88,-0.01) node{\footnotesize Noise};
    \end{scope}
    \end{tikzpicture}

    \caption{ Image restoration results of GPT-Image on real-world degraded images without ground truth. Each vertical pair shows a degraded input image (top) and its corresponding restored output (bottom), with the type of degradation labeled beside each pair.}
    \label{fig:fig2}
\end{figure*}

\begin{figure*}[!t]
    \centering

    \begin{tikzpicture}
    \centering
    \node[anchor=south west,inner sep=0] (image) at (0,0) {\includegraphics[width=0.99\linewidth]{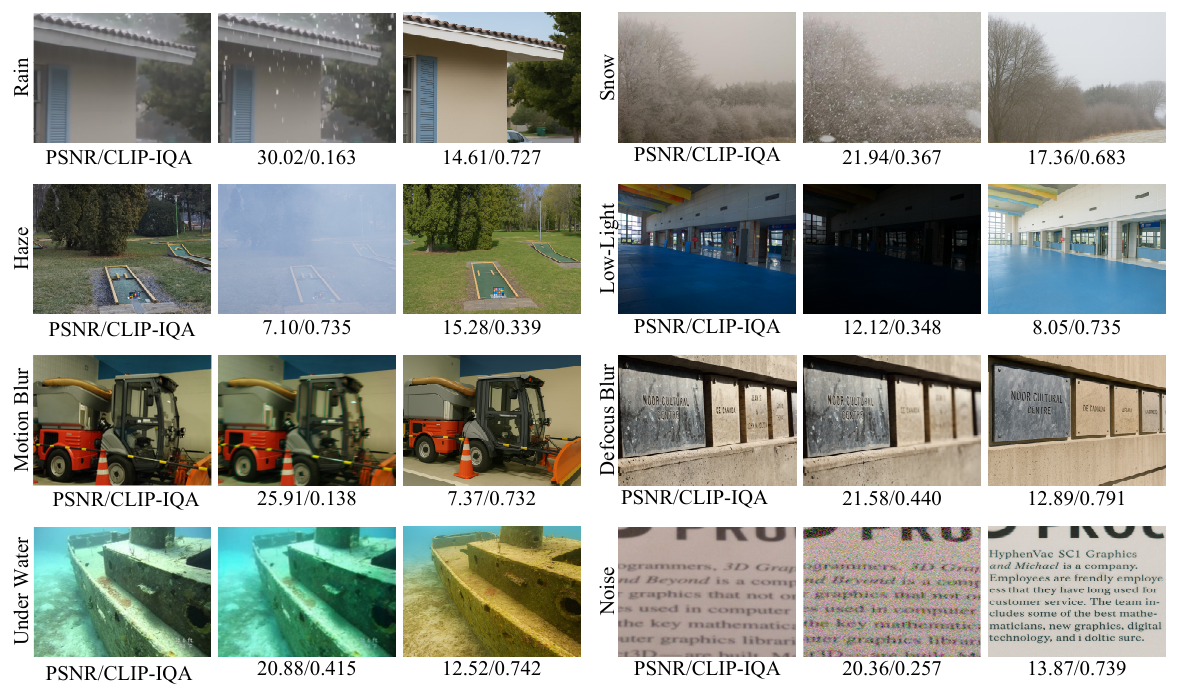}};
    \begin{scope}[x={(image.south east)},y={(image.north west)}]

    \end{scope}
    \end{tikzpicture}

    \caption{ Image restoration results of GPT-Image on real-world degraded images with available ground truth. Each triplet consists of the ground truth image, the degraded input, and the corresponding restored output, with the type of degradation labeled beside each set. We display the PSNR and CLIP-IQA scores below each image, reflecting perceptual quality and pixel-level structural fidelity, respectively.
}
    \label{fig:fig2-2}

\end{figure*}

(i) \textbf{Restoration Capability of GPT-Image}: We evaluate GPT-Image on eight diverse image restoration tasks through both quantitative and qualitative analysis. While the restored images are visually appealing (as reflected by CLIP-IQA \cite{wang2023exploring} scores), they often suffer from a lack of pixel-level structural fidelity, as indicated by lower PSNR scores even compared with the degraded image (e.g., 12.89 dB vs. 21.58 dB).

(ii) \textbf{Failure Cases}: Although GPT-Image generally preserves overall image semantics, it often fails to maintain pixel-level structural fidelity. This is primarily due to three limitations: distortion of image proportions, inaccuracies in object positioning and quantity, and inconsistencies in viewpoint reconstruction, which are often critical for low-level image restoration tasks.

(iii) \textbf{A Baseline}: Although GPT-Image performs poorly in preserving pixel-level structural fidelity, its visually pleasing outputs can serve as strong priors. We propose a lightweight post-processing baseline that leverages GPT-Image's outputs to enhance the performance of image restorations.

\section{Related Work}
\noindent\textbf{Image Restoration.} Image restoration \cite{yang2024language} aims to reconstruct high-quality images from inputs degraded by diverse factors such as rain \cite{zhang2019image_rain800}, snow \cite{liu2018desnownet}, haze \cite{ancuti2018haze}, low-light \cite{Chen2018Retinex}, motion blur \cite{rim2020real}, defocus blur \cite{abuolaim2020defocus}, underwater distortion \cite{wang2024underwater}, and noise \cite{Buades2005NonLocal}. Early traditional methods rely on handcrafted priors (e.g., dark channel prior \cite{he2010single} for haze removal or bilateral filtering \cite{papari2016fast} for denoising), but often failed under real-world complexity. With the rise of deep learning, task-specific restoration networks have emerged. For instance, haze removal benefits from spatial priors and transformer-based models~\cite{song2023vision}, while rain and snow removal have been addressed through multi-scale CNNs~\cite{jiang2020multi}. Low-light enhancement leverages illumination-aware representations~\cite{jiang2021enlighte}, and underwater image enhancement exploits color correction and local contrast adaptation~\cite{wang2022underwater}. Blur-related degradations (motion and defocus) are often handled with deblurring networks that preserve spatial structure~\cite{zamir2022restormer}. For noise, residual learning-based denoisers like DnCNN remain effective baselines~\cite{zhang2017dncnn}.
Recently, universal image restoration has been explored using generative priors, particularly diffusion-based \cite{ozdenizci2023restoring} or vision-language models \cite{liu2025up}, which offer semantic-level guidance and strong generalization across degradation types.

\begin{figure*}[!t]
    \centering

    \begin{tikzpicture}
    \centering
    \node[anchor=south west,inner sep=0] (image) at (0,0) {\includegraphics[width=0.99\linewidth]{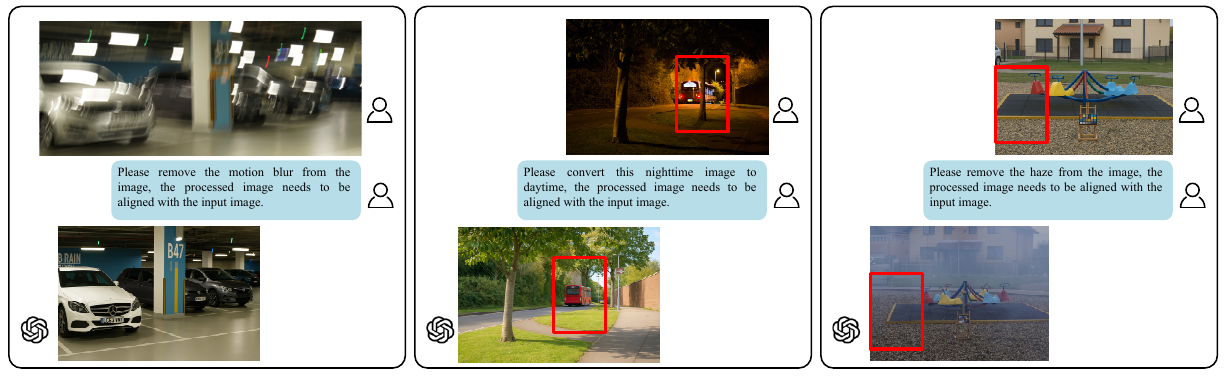}};
    \begin{scope}[x={(image.south east)},y={(image.north west)}]

    \draw (0.17,-0.02) node{\footnotesize (a)};
    \draw (0.5,-0.02) node{\footnotesize (b)};
    \draw (0.84,-0.02) node{\footnotesize (c)};
    
    \end{scope}
    \end{tikzpicture}

    \caption{Failure cases. (a) Variations in image proportions. (b) Shifts in object positions and quantities. (c) Changes in viewpoint.}
    \label{fig:fig3}

\end{figure*}


\noindent\textbf{Text-guided Image Editing.} Text-guided image editing has become a central topic in generative visual manipulation, aiming to modify an existing image according to natural language instructions while preserving irrelevant regions. Early approaches such as DiffusionCLIP~\cite{diffusionclip2023} and Null-Text Inversion~\cite{nulltext2022} leverage pretrained diffusion models and CLIP embeddings to apply prompt-driven semantic edits without explicit supervision. InstructPix2Pix~\cite{instructpix2pix2023} introduce instruction tuning into diffusion models by constructing synthetic image-instruction pairs, enabling edit operations like ``make it sunset'' or ``remove the hat'' via a single prompt. To enhance edit fidelity and reconstruction consistency, subsequent works such as Qwen \cite{wu2025qwen} propose improved inversion strategies and sampling schemes that better preserve image structure while achieving prompt compliance.

To broaden controllability and usability, recent methods integrate human feedback, spatial priors, or large multimodal language models. For example, DragDiffusion~\cite{dragdiffusion2024} enables fine-grained point-based dragging, while MeshPad~\cite{li2025meshpad} performs sketch-conditioned inpainting. Vitron~\cite{fei2024vitron} employs multimodal instruction tuning with vision-language models to interpret user intent and support multi-turn, task-specific editing. Models such as ReferDiffusion~\cite{liu2024referring} further fuse segmentation or audio cues with textual prompts, pushing toward general-purpose, instruction-following editing. These advances signify a shift from simple prompt modulation to rich, multimodal, user-centric control. However, they remain difficult to apply directly to image restoration. This work explores integrating such models, particularly GPT-Image, as informative priors to enhance restoration performance.

\begin{figure*}[!t]
    \centering

    \begin{tikzpicture}
    \centering
    \node[anchor=south west,inner sep=0] (image) at (0,0) {\includegraphics[width=0.85\linewidth]{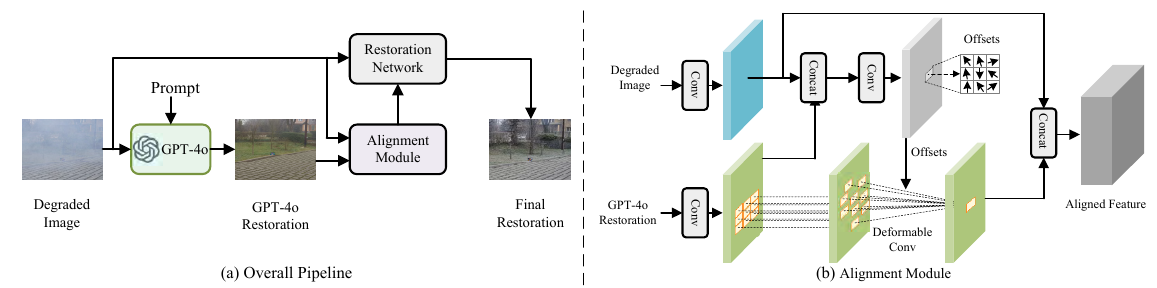}};
    \begin{scope}[x={(image.south east)},y={(image.north west)}]

    \end{scope}
    \end{tikzpicture}

    \caption{Pipeline of our proposed solution. (a) Overall Pipeline, (b) the structure of Alignment Module. We use GPT-Image-generated images as priors, align them with the degraded image using a deformable convolution-based alignment module, and feed the aligned features together with the degraded image into a restoration backbone to obtain the final restored result.
}
    \label{fig:method1}

\end{figure*}

\section{Restoration of Diverse Degradation}

\cref{fig:fig2} and \cref{fig:fig2-2} present the restoration results of GPT-Image on eight representative types of real-world degradation. The degraded images are collected from datasets related to deraining \cite{fu2019lightweight}, desnowing \cite{liu2018desnownet}, dehazing \cite{ancuti2018haze}, low-light enhancement \cite{liu2024ntire}, motion deblurring \cite{rim2020real}, defocus deblurring \cite{yang2024ldp}, underwater image enhancement \cite{li2019underwater}, and denoising \cite{liu2024residual}, as well as from web sources. For the real-world images with ground truth shown in \cref{fig:fig2-2}, we report quantitative metrics including PSNR and CLIP-IQA \cite{wang2023exploring}. The first metric evaluates pixel-level structural fidelity, while the latter two assess perceptual quality.

Overall, GPT-Image delivers visually compelling restorations across a wide range of image restoration tasks, showcasing its versatility. For example, in deraining and desnowing, it effectively removes occlusions like rain streaks and snow buildup, restoring clean scenes with preserved fine details in trees, pedestrians, and vehicles. These results highlight GPT-Image’s potential not only in task-specific restoration but also as a unified foundation model for general-purpose low-level vision restoration.

However, as shown in \cref{fig:fig2-2}, while GPT-Image achieves high CLIP-IQA scores (indicating strong perceptual quality), its PSNR values are often lower than even those of the degraded input. This reveals a significant limitation: poor preservation of pixel-level structural fidelity, which is critical for many practical restoration applications.

\section{Failure Cases}
We analyze several representative cases to further investigate the pixel-level structural fidelity issues present in GPT-Image’s restoration results.

\noindent\textbf{Variations in Image Proportions.}
As shown in the left part of \cref{fig:fig3}, GPT-Image fails to preserve the original aspect ratio during restoration, leading to noticeable geometric distortions. Such inconsistencies disrupt visual coherence and can be detrimental to downstream tasks that depend on accurate spatial representation.

\noindent\textbf{Shifts in Object Positions and Quantities.}
In the middle example of \cref{fig:fig3}, GPT-Image exhibits poor control over object presence and placement. For instance, it inadvertently removes a roadside tree, despite no instruction to modify the scene content. This highlights a key challenge in maintaining structural and semantic consistency for image restoration within multimodal generation frameworks.

\noindent\textbf{Changes in Viewpoint.}
On the right side of \cref{fig:fig3}, GPT-Image applies slight scaling and cropping, altering the original camera viewpoint. As a result, certain scene elements, such as a swing set in the lower-left corner, are partially or entirely lost. Such viewpoint shifts can undermine restoration reliability, especially when precise scene reconstruction is required.

While GPT-Image demonstrates impressive generative capabilities and generalization across diverse image restoration tasks, it exhibits notable limitations in maintaining geometric consistency, accurate object placement, and stable viewpoints for achieving high pixel-level structure fidelity. These shortcomings can be critical in applications where spatial precision is essential. Addressing them will be vital for advancing the reliability of multimodal models in image restoration tasks.

\begin{table*}[!t]
\centering
\caption{Quantitative results on O-Haze (dehazing), Rain800 (deraining), and LOL (low-light enhancement) datasets.}

\setlength{\tabcolsep}{0.8pt}
\renewcommand{\arraystretch}{1}
\begin{tabular}{lccccccccc}
\toprule
\multirow{2}{*}{Method} 
& \multicolumn{3}{c}{O-Haze \cite{ancuti2018haze}} 
& \multicolumn{3}{c}{Rain800 \cite{zhang2019image_rain800}} 
& \multicolumn{3}{c}{LOL \cite{Chen2018Retinex}} \\
\cmidrule(lr){2-4} \cmidrule(lr){5-7} \cmidrule(lr){8-10}
& PSNR$\uparrow$ & SSIM$\uparrow$ & CLIP-IQA$\uparrow$
& PSNR$\uparrow$ & SSIM$\uparrow$ & CLIP-IQA$\uparrow$
& PSNR$\uparrow$ & SSIM$\uparrow$ & CLIP-IQA$\uparrow$ \\
\midrule
GPT-Image \cite{openai_gpt_image_1}       
& 13.13 & 0.133 & 0.757  
& 12.44 & 0.296 & 0.812  
& 12.13 & 0.387 & 0.706  \\
Baseline \cite{zamir2022restormer} 
& 20.86 & 0.794 & 0.540  
& 28.63 & 0.881 & 0.612  
& 21.28 & 0.807 & 0.470  \\
Ours                                 
& 22.08 & 0.801 & 0.566  
& 29.19 & 0.893 & 0.628  
& 22.18 & 0.831 & 0.495  \\
\bottomrule
\end{tabular}
\label{tab:3datasets-A}

\end{table*}

\begin{table*}[!t]
\centering
\caption{Quantitative results on RainDrop (raindrop removal), Nature20 (reflection removal), and UIEB (underwater enhancement) datasets.}

\setlength{\tabcolsep}{0.8pt}
\renewcommand{\arraystretch}{1}
\begin{tabular}{lccccccccc}
\toprule
\multirow{2}{*}{Method} 
& \multicolumn{3}{c}{RainDrop \cite{qian2018attentive}} 
& \multicolumn{3}{c}{Nature20 \cite{li2020single}} 
& \multicolumn{3}{c}{UIEB \cite{li2019underwater}} \\
\cmidrule(lr){2-4} \cmidrule(lr){5-7} \cmidrule(lr){8-10}
& PSNR$\uparrow$ & SSIM$\uparrow$ & CLIP-IQA$\uparrow$
& PSNR$\uparrow$ & SSIM$\uparrow$ & CLIP-IQA$\uparrow$
& PSNR$\uparrow$ & SSIM$\uparrow$ & CLIP-IQA$\uparrow$ \\
\midrule
GPT-Image \cite{openai_gpt_image_1}       
& 15.73 & 0.404 & 0.691 
& 14.72 & 0.456 & 0.700 
& 11.88 & 0.313 & 0.785 \\
Baseline \cite{zamir2022restormer} 
& 30.07 & 0.911 & 0.418 
& 23.80 & 0.818 & 0.402 
& 21.67 & 0.893 & 0.451 \\
Ours                                 
& 30.53 & 0.914 & 0.420 
& 24.72 & 0.823 & 0.415 
& 21.95 & 0.899 & 0.456 \\
\bottomrule
\end{tabular}
\label{tab:3datasets-B}

\end{table*}

\section{A Baseline Solution}
To mitigate the aforementioned limitations, we propose using the image restored by GPT-Image as a powerful prior to further improve image restoration performance. We take image dehazing, deraining, low-light enhancement, raindrop removal, reflection removal, and underwater enhancement as test cases and explore a baseline network, as a plug-in-and-play model, that post-processes GPT-Image’s restoration outputs to improve pixel-level structural fidelity.

\noindent\textbf{Overall Pipeline.} As shown in \cref{fig:method1}, given a degraded image, we first employ GPT-Image with a task-specific prompt to generate an initial restoration, referred to as GPT-Image Restoration. This serves as a strong prior to guide the subsequent restoration process. To address potential misalignment between the degraded input and the GPT-Image output, both are fed into an Alignment module, which aligns structural content from the two sources. The aligned features are then processed by a Restoration Network to produce the final high-quality output. This collaborative pipeline leverages GPT-Image as an external prior, providing a simple yet effective means to enhance image restoration performance. The alignment module and restoration network are based on the DCN \cite{zhu2019deformable} and Restormer \cite{zamir2022restormer}, respectively. The prompt used to instruct GPT-Image for image restoration is: [\textit{Please remove the \{degradation type\} from the image. The processed image should remain aligned with the input image.}]

\noindent\textbf{Implementation Details.}  The network is trained using the charbonnier loss \cite{jiang2020multi} for the O-Haze dataset \cite{ancuti2018haze} (40 training and 5 testing images), Rain800 dataset \cite{zhang2019image_rain800}  (700 training and 100 testing images), LOL \cite{Chen2018Retinex} (485 training and 15 testing images), RainDrop \cite{qian2018attentive} (861 training and 58 testing images), Nature20 \cite{li2020single} (200 training and 20 testing images) and UIEB \cite{li2019underwater} (700 training and 109 testing). All experiments are conducted using NVIDIA RTX 4090 and implemented in PyTorch. Training is performed using the Adam optimizer with an initial learning rate of $2\times10^{-4}$, decayed via a cosine annealing schedule. We use a batch size of 2, and input images are randomly cropped into $256 \times 256$ patches. Standard data augmentation techniques, including random horizontal flipping and random rotation, are applied. The network is trained for a total of 150,000 iterations.

\noindent\textbf{Metric.} We use Peak Signal-to-Noise Ratio (PSNR) and Structural Similarity (SSIM) to evaluate pixel-wise image fidelity, and CLIP-IQA \cite{wang2023exploring} to assess perceptual image quality.

\begin{figure*}[!t]
    \centering

    \begin{tikzpicture}
    \centering
    \node[anchor=south west,inner sep=0] (image) at (0,0) {\includegraphics[width=0.99\linewidth]{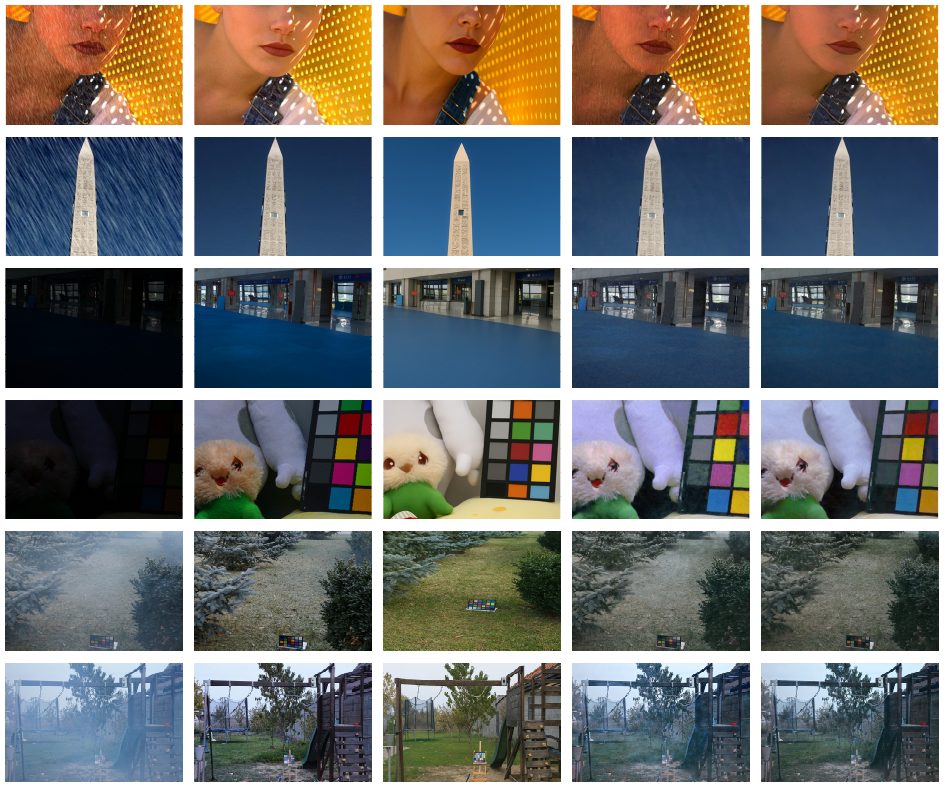}};
    \begin{scope}[x={(image.south east)},y={(image.north west)}]

    \draw (0.10,-0.01) node{\footnotesize Degraded};
    \draw (0.30,-0.01) node{\footnotesize Ground Truth};
    \draw (0.50,-0.01) node{\footnotesize GPT-Image };
    \draw (0.70,-0.01) node{\footnotesize Baseline};
    \draw (0.91,-0.01) node{\footnotesize Ours};

    \end{scope}
    \end{tikzpicture}

    \caption{Comparisons on the Rain800 \cite{zhang2019image_rain800}, LOL \cite{Chen2018Retinex}, and O-HAZE \cite{ancuti2018haze} datasets. Rows 1–2 show results on Rain800 dataset, Rows 3–4 are results on LOL dataset, and Rows 5–6 are for O-HAZE dataset. GPT-Image  denotes the image restoration results generated by GPT-Image . Baseline refers to the restoration results without using GPT-Image  priors, while Ours indicates the enhanced restoration results guided by GPT-Image  priors.}
    \label{fig:fig5}

\end{figure*}

\begin{figure*}[!t]
    \centering

    \begin{tikzpicture}
    \centering
    \node[anchor=south west,inner sep=0] (image) at (0,0) {\includegraphics[width=0.99\linewidth]{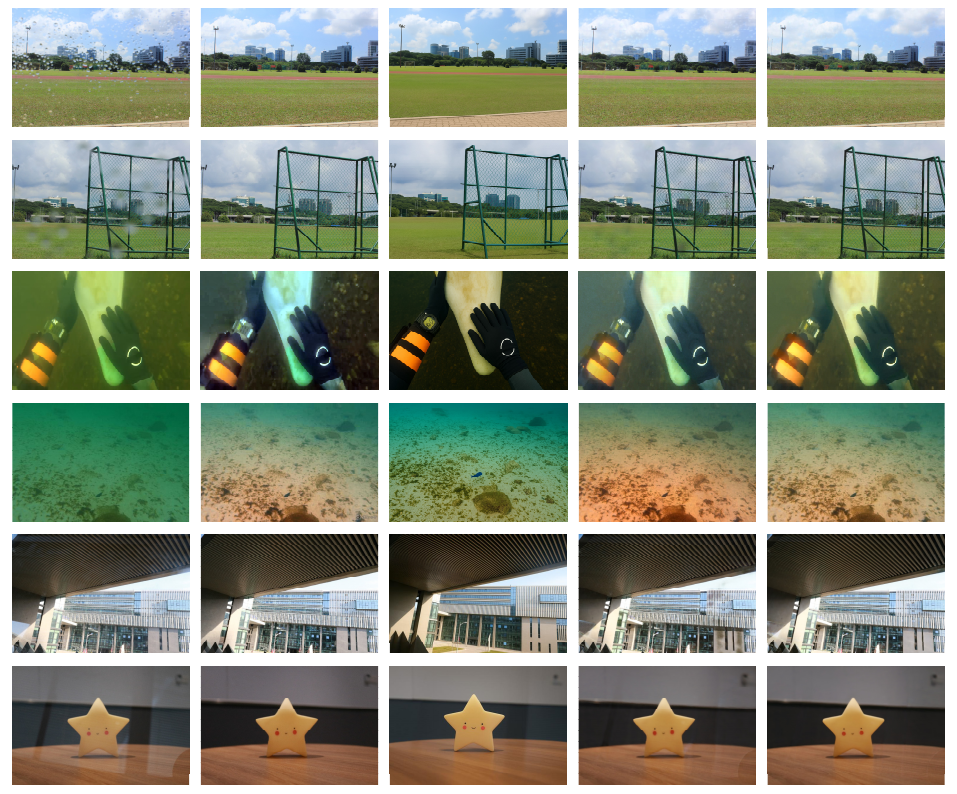}};
    \begin{scope}[x={(image.south east)},y={(image.north west)}]

    \draw (0.10,-0.01) node{\footnotesize Degraded};
    \draw (0.30,-0.01) node{\footnotesize Ground Truth};
    \draw (0.50,-0.01) node{\footnotesize GPT-Image };
    \draw (0.70,-0.01) node{\footnotesize Baseline};
    \draw (0.91,-0.01) node{\footnotesize Ours};

    \end{scope}
    \end{tikzpicture}

    \caption{Comparisons on the RainDrop \cite{qian2018attentive}, UIEB \cite{li2019underwater}, and Nature20 \cite{li2020single} datasets. Rows 1–2 show results on RainDrop dataset, Rows 3–4 are results on UIEB dataset, and Rows 5–6 are for Nature20 dataset. GPT-Image  denotes the image restoration results generated by GPT-Image . Baseline refers to the restoration results without using GPT-Image  priors, while Ours indicates the enhanced restoration results guided by GPT-Image  priors.}
    \label{fig:figNature20}

\end{figure*}

\begin{table}[!t]
    \centering
    \caption{Quantitative Results on the UIEB dataset for different backbones, with and without GPT-Image priors. T, C, and M denote neural network architectures based on Transformer, CNN, and Mamba, respectively.}

    \setlength{\tabcolsep}{1pt}
    \renewcommand{\arraystretch}{1}
    \begin{tabular}{lccccccc}
        \toprule
        \multirow{2}{*}{Backbone} & \multirow{2}{*}{Type} 
        & \multicolumn{3}{c}{w/o GPT-Image} 
        & \multicolumn{3}{c}{w/ GPT-Image (Ours)} \\
        \cmidrule(lr){3-5} \cmidrule(lr){6-8}
        & & PSNR$\uparrow$ & SSIM$\uparrow$ & CLIP-IQA$\uparrow$
          & PSNR$\uparrow$ & SSIM$\uparrow$ & CLIP-IQA$\uparrow$ \\
        \midrule
        Restormer      & T   &21.67  &0.893  &0.451  &21.95  &0.899  &0.456\\
        ConvIR         & C           &22.23 &0.903 &0.414  &22.75 &0.904 &0.433 \\
        X-Restormer    & T   & 22.04 & 0.897 & 0.395  & 22.74 & 0.908 & 0.440 \\
        MambaIRv2      & M           &22.40 &0.908 &0.436  &22.91 &0.913 &0.461 \\
        \bottomrule
    \end{tabular}
    \label{tab:backbone_gpt4o_ohaze}

\end{table}

\begin{table}[!t]
\centering
\caption{Effectiveness of the Alignment module.}

\setlength{\tabcolsep}{11pt}
\begin{tabular}{lccc}
\Xhline{2\arrayrulewidth}
Fusion strategy & PSNR$_\uparrow$ & SSIM$_\uparrow$ & CLIP-IQA$_\uparrow$ \\
\Xhline{1\arrayrulewidth}
Baseline &21.67 &0.893 &0.451 \\

Concat &21.75 &0.895 &0.450  \\

Ours (Alignment module) 
&21.95 &0.899 &0.456\\
\Xhline{2\arrayrulewidth}
\end{tabular}
\label{tab:abs}

\end{table}

\noindent\textbf{Results.} We compare two baselines: (i) the direct restoration output from GPT-Image, and (ii) a standard Restormer model trained to restore directly from the degraded image. Quantitative results are presented in \cref{tab:3datasets-A} and \cref{tab:3datasets-B}. Our pipeline using GPT-Image outputs as visual priors achieves significantly higher scores in perceptual quality metrics (\text{e.g.,} 0.566 in CLIP-IQA on the O-Haze dataset), indicating improved visual appeal. At the same time on the O-Haze dataset, it achieves comparable performance in pixel-level structural metrics (e.g., 22.08 in PSNR), demonstrating that the enhancement in visual quality does not come at the expense of structural fidelity.

We present a visual comparison in \cref{fig:fig5} and \cref{fig:figNature20}. The first column shows the degraded input images, the fourth column displays the restoration results from the baseline Restormer (without GPT-Image guidance), and the last column presents the outputs of our proposed method incorporating aligned GPT-Image priors. Across a variety of challenging scenes, our method consistently produces clearer restorations with reduced artifacts (noise) compared to the baseline. For example, in the outdoor slide scene, our approach successfully recovers fine details in the slide, whereas the baseline result appears desaturated and lacks contrast. Similarly, in the forest pathway scene, our method restores distant foliage and pathway textures with enhanced sharpness and color fidelity. Consistent improvements are also observed on deraining, low-light enhancement and others, further demonstrating the effectiveness of our method. These improvements highlight the effectiveness of integrating GPT-Image-generated priors to enhance restoration quality.

\noindent\textbf{Generality of GPT-Image Priors.} To further validate the generality of this prior, we extend our experiments beyond Restormer \cite{zamir2022restormer} to other baselines, including ConvIR \cite{cui2024revitalizing}, X-Restormer \cite{chen2024comparative}, and MambaIRv2 \cite{guo2024mambairv2}. As shown in Tab. \ref{tab:backbone_gpt4o_ohaze}, the proposed pipeline incorporating GPT-Image outputs as visual priors achieves significantly better scores on perceptual quality metrics (e.g., CLIP-IQA on UIEB dataset increases by 0.025 on MambaIRv2.), indicating improved perceptual fidelity. At the same time, it maintains competitive performance on pixel-level structural metrics on the UIEB dataset (e.g., PSNR of 22.91dB), suggesting that the perceptual enhancement does not come at the cost of structural integrity. Even compared to the state-of-the-art image restoration method, MambaIRv2 (22.40dB PSNR), our framework (22.91dB PSNR) achieves superior performance, indicating the superiority of our method. These results confirm that the GPT-Image priors consistently improve performance across CNN, Transformer, and Mamba-based restoration backbones. 

\noindent\textbf{Effectiveness of the Alignment module.}
To validate the effectiveness of the proposed Alignment module we conduct an ablation study in Table \cref{tab:abs}. As shown, replacing our fusion module with a simple concatenation yields only marginal improvements over the baseline. In contrast, incorporating our Alignment module consistently delivers the best performance.

\begin{figure}[!t]
    \centering
    \begin{tikzpicture}
        \node[anchor=south west,inner sep=0] (image) at (0,0) {
            \includegraphics[width=0.99\textwidth]{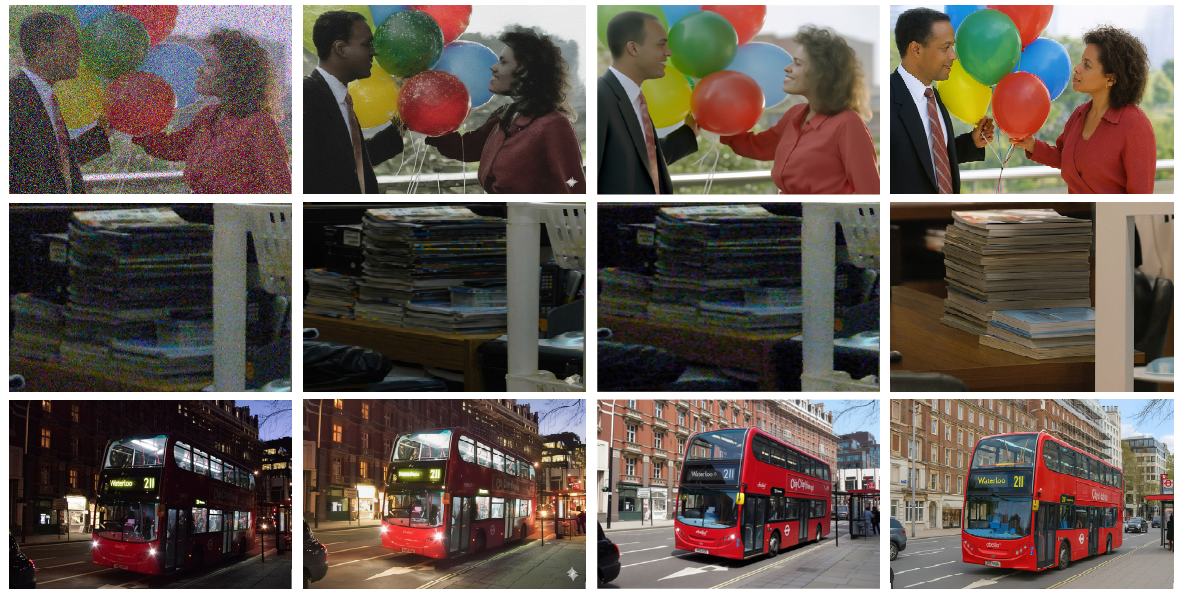}
        };
        \begin{scope}[x={(image.south east)}, y={(image.north west)}]
            \draw (0.1,-0.05) node {\footnotesize Degraded};
            \draw (0.375,-0.05) node {\footnotesize Nano Banana Pro};
            \draw (0.62,-0.05) node {\footnotesize Qwen3};
            \draw (0.88,-0.05) node {\footnotesize GPT-Image};
        \end{scope}
    \end{tikzpicture}

    \caption{Typical image-editing models for image restoration tasks include Nano Banana Pro, Qwen3, and GPT-Image.}
    \label{fig:fig6}

\end{figure}

\section{Discussion}
We further compare GPT-Image with two other state-of-the-art multimodal models, Nano Banana Pro (Gemini 3) \cite{team2023gemini} and Qwen3 \cite{wu2025qwen}, in terms of image restoration performance, as shown in Fig. 6. GPT-Image consistently delivers more stable, sharper, and structurally realistic restoration results than Nano Banana Pro and Qwen3. Notably, GPT-Image better preserves fine-grained details such as subtle object boundaries and texture continuity, whereas Nano Banana Pro and Qwen3 sometimes introduce artifacts or overly smooth delicate structures in the scene. These observations indicate that GPT-Image currently provides superior visual fidelity for restoration-oriented generative tasks. This performance advantage remains consistent across diverse image contents and degradation conditions, demonstrating GPT-Image’s robustness in various scenarios. However, all three models exhibit slight pixel-level misalignment, further highlighting the need for alignment mechanisms when integrating generative priors into low-level vision pipelines. In addition, there is a significant difference in computational efficiency: GPT-Image requires an average of 82 seconds per image, whereas Nano Banana Pro and Qwen3 take only 27 seconds and 18 seconds, respectively. This underscores the practical trade-off between restoration quality and inference speed during real-world deployment.

\section{Conclusion}

In this study, We present the first systematic evaluation of GPT-Image for image restoration across diverse degradations. While GPT-Image excels in generating perceptually pleasing results, it often lacks pixel-level structural fidelity, exhibiting geometric distortions and object misalignments. We show that GPT-Image outputs can serve as strong visual priors when combined with a lightweight post-processing network, effectively enhancing structural accuracy without sacrificing visual quality. Our findings highlight the potential of leveraging large multimodal models for restoration and offer guidance for future research in this direction.

\bibliographystyle{abbrvnat}  
\small
\bibliography{Reference}

\end{document}